\documentclass[11pt]{article}

\usepackage[margin=2.5cm]{geometry}
\usepackage{times}
\usepackage{amsmath,amssymb}
\usepackage{booktabs}
\usepackage{graphicx}
\usepackage{float}
\usepackage{url}
\usepackage[breaklinks,colorlinks,citecolor=blue,urlcolor=blue,linkcolor=black]{hyperref}
\usepackage[numbers,sort&compress]{natbib}
\usepackage{caption}

\title{Answer-Conditioned Chain-of-Thought Distillation for\\Few-Shot Industrial Vision with Small VLMs}

\author{
  Shubham Rao\\
  Entropy AI Research Labs Private Limited\\
  \texttt{director@entropyresearch.ai}\\
  \url{https://entropyresearch.ai}
}

\date{}

\begin{document}
\maketitle

\begin{abstract}
Deploying AI-based visual inspection in manufacturing is hard because requirements change often, new defect types appear, and large labeled datasets are rarely available. We propose answer-conditioned chain-of-thought (CoT) distillation for rapidly adapting small vision-language models (VLMs) to new industrial tasks using minimal labeled data. A frontier VLM receives each training image along with its correct label and generates a justified visual explanation. A 3B-parameter model is then fine-tuned on these reasoning-augmented examples via LoRA. By conditioning on correct answers, we ensure all training reasoning is directed toward the correct conclusion, which is critical because frontier models score as low as 24.1\% on our hardest task. We validate on four industrial classification tasks spanning three image modalities using only 18 to 30 labeled images per task. Across 4 seeds per task (32 training runs), our method outperforms direct fine-tuning on all 16 seed-task combinations, with mean improvements of +1.7 to +4.4 percentage points. A controlled equal-budget experiment confirms the improvement comes from reasoning quality, not additional training steps. An unconditioned baseline demonstrates that without answer-conditioning, wrong reasoning degrades performance by 17.8 percentage points. On weld radiograph classification, the fine-tuned 3B model outperforms GPT-4.1 by 10.0pp using just 24 training images.
\end{abstract}

\section{Introduction}
\label{sec:intro}

Visual quality inspection is central to manufacturing. Every production line needs to identify defects, classify materials, or verify compliance with standards. These requirements change often. Each change demands a new or retrained vision model.

Traditional approaches rely on CNNs trained on thousands of labeled images~\cite{czimmermann2020visual}. Collecting this data is expensive. In many settings, a factory has fewer than 30 labeled examples and needs a working classifier within days, not months.

Large VLMs like GPT-4.1 can interpret images without task-specific training. But their accuracy on specialized industrial tasks is poor. On concrete aggregate grading (9 classes, DIN 1045~\cite{din1045}), GPT-4.1 achieves 29.6\% few-shot. Gemini 2.5 Pro achieves 24.1\%. These models cannot be deployed on-premises due to cost and data privacy constraints.

Small VLMs (1--7B parameters)~\cite{wang2024qwen2vl} fit on edge hardware and can be fine-tuned with LoRA~\cite{hu2021lora}. But standard fine-tuning on 18--30 images maps images to labels without transferring domain knowledge. The model memorizes rather than learns.

CoT distillation~\cite{hsieh2023distilling} trains smaller models on reasoning-augmented data from larger models, but existing work targets text-only settings. STaR~\cite{zelikman2022star} and Video-STaR~\cite{zohar2025videostar} use answer-conditioned reasoning in text and video settings, but not for VLMs where the teacher itself performs poorly.

We propose answer-conditioned CoT distillation for VLMs. A frontier model receives each training image with its correct label and explains why that label is correct. A small VLM is fine-tuned on these reasoning-augmented pairs via LoRA. The frontier model is not classifying. It already knows the answer. It is explaining what visual features justify that answer.

This is critical because frontier model accuracy ranges from 24.1\% to 91.1\% across our tasks. If we asked it to classify and explain, up to 76\% of training data would contain incorrect reasoning. We show experimentally that unconditioned CoT destroys performance ($-$17.8pp) when the teacher is mostly wrong.

Our contributions:
\begin{enumerate}
\item We apply answer-conditioned CoT distillation to VLMs for industrial few-shot classification, showing consistent improvement across 4 tasks, 3 modalities, and 4 random seeds (16/16 wins).
\item A controlled equal-budget experiment demonstrates that the improvement comes from reasoning quality, not additional training steps.
\item We show that without answer-conditioning, unconditioned CoT degrades performance by up to 17.8pp.
\item A 3B model fine-tuned on 24 images outperforms GPT-4.1 and Gemini 2.5 Pro on specialized industrial tasks.
\end{enumerate}

\section{Related Work}
\label{sec:related}

\textbf{Chain-of-thought distillation.} CoT prompting~\cite{wei2022chain} improves performance by including reasoning steps. Hsieh et al.~\cite{hsieh2023distilling} showed training small LMs on reasoning-augmented outputs outperforms standard fine-tuning. Their work is text-only. We extend this to VLMs with image-grounded visual reasoning.

\textbf{Answer-conditioned reasoning.} STaR~\cite{zelikman2022star} bootstraps reasoning by filtering for correct answers. Rajani et al.~\cite{rajani2019explain} generate explanations conditioned on correct answers for commonsense reasoning. Video-STaR~\cite{zohar2025videostar} applies self-training with answer conditioning to video. All operate where the teacher is already competent. We apply answer-conditioning where teacher accuracy is as low as 24.1\%.

\textbf{Knowledge distillation for VLMs.} VLsI~\cite{vlsi2024verbalized} distills internal representations. KD for long-tail recognition~\cite{kd2024longtail} uses logit-based distillation. Online In-Context Distillation~\cite{online2025incontext} maintains the teacher at inference time. None use answer-conditioned reasoning, and none target industrial few-shot settings.

\textbf{Industrial vision with VLMs.} Adapting foundation models for industry~\cite{adapting2024industry} typically requires large datasets. CLIP-based few-shot approaches~\cite{clip2025fewshot} require 50--100 examples. Xia et al.~\cite{xia2025bootstrapping} bootstrap grounded CoT in multimodal LLMs but do not condition on correct answers.

\section{Method}
\label{sec:method}

\subsection{Prompt Design}

Each task uses a single classification prompt containing: (1) image context, (2) class definitions with visual descriptions, (3) contrastive features distinguishing similar classes, and (4) JSON output format. The same prompt is used for training and evaluation.

Prompt quality is critical. In early experiments on steel surface defects, using shorter class definitions during training (while keeping the full definitions for evaluation) degraded accuracy by 18.1 percentage points (48.6\% vs 66.7\%), with the same training images and hyperparameters. The class definitions must be detailed and identical between training and evaluation.

\subsection{Answer-Conditioned CoT Generation}

For each training image $(x_i, y_i)$, the frontier model (GPT-4.1) receives the correct label $y_i$ with the image $x_i$. It explains why $y_i$ is correct using contrastive reasoning. We generate 3 descriptions per image at temperature 0.7.

An example generated description for a weld radiograph classified as \textit{porosity}:

\begin{quote}
\small\textit{``The image shows several scattered small dark circular spots distributed across the weld area. These spots are characteristically round, consistent with gas pores trapped during solidification. This is not lack\_of\_penetration, which would show a continuous dark line along the weld centerline, nor cracks, which would appear as sharp jagged lines.''}
\end{quote}

The frontier model is not classifying. It already knows the answer. It is explaining what visual features justify that answer and why similar classes do not apply. The JSON output is never generated by the frontier model; it is appended programmatically to guarantee format consistency.

\subsection{Training Data Construction}

From the same $N$ labeled images, we construct:

\textbf{Direct LoRA:} $N$ pairs (image + prompt $\rightarrow$ JSON label).

\textbf{CoT-Augmented LoRA:} $4N$ pairs (3 CoT + 1 direct per image). For CoT pairs, the ``Respond with JSON'' instruction is stripped from the user prompt. The assistant response contains the description followed by the correct JSON.

\begin{figure}[H]
\centering
\includegraphics[width=\textwidth]{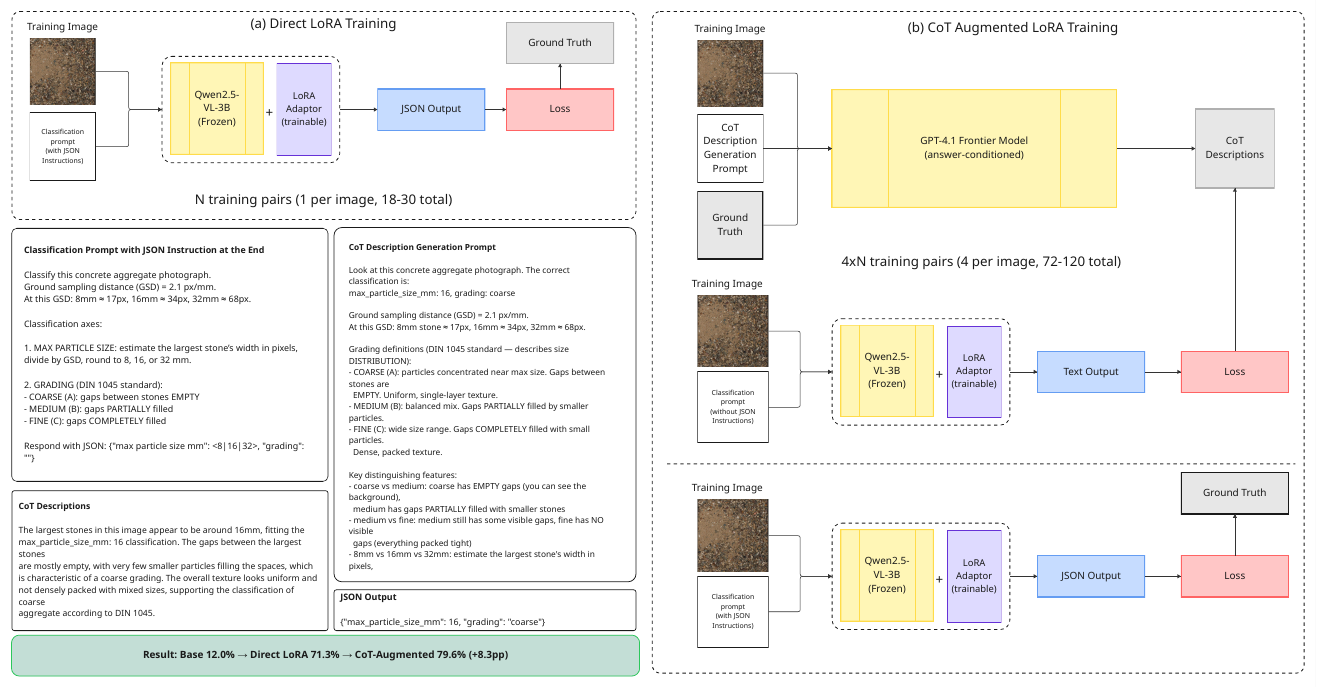}
\caption{Method overview. (a) Direct LoRA learns image$\rightarrow$JSON from $N$ pairs. (b) CoT-augmented LoRA: GPT-4.1 receives each image with the correct label and generates justified descriptions, yielding $4N$ training examples from $N$ images.}
\label{fig:pipeline}
\end{figure}

Figure~\ref{fig:pipeline} illustrates both approaches.

\subsection{LoRA Fine-Tuning}

Both approaches use identical hyperparameters on Qwen2.5-VL-3B-Instruct~\cite{wang2024qwen2vl} (Table~\ref{tab:hyperparams}). Only the assistant response tokens contribute to the loss; all user-prompt and image tokens are masked.

\begin{table}[h]
\centering\small
\begin{tabular}{ll}
\toprule
Parameter & Value \\
\midrule
Base model & Qwen2.5-VL-3B-Instruct (3.8B params) \\
Precision & BF16 (no quantization) \\
LoRA rank / alpha / dropout & 16 / 32 / 0.05 \\
Target modules & q, k, v, o, gate, up, down proj \\
Trainable parameters & 37.2M (0.98\% of total) \\
Learning rate & $2 \times 10^{-5}$ \\
Epochs & 40 \\
Effective batch size & 4 (gradient accumulation) \\
Scheduler & Cosine with 10\% warmup \\
\bottomrule
\end{tabular}
\caption{Training hyperparameters (identical for Direct and CoT-Aug).}
\label{tab:hyperparams}
\end{table}

\section{Experimental Setup}
\label{sec:experiments}

\subsection{Datasets}

We evaluate on four tasks spanning three image modalities (Table~\ref{tab:datasets}).

\begin{table}[h]
\centering\small
\begin{tabular}{llccc}
\toprule
Dataset & Modality & Classes & Train & Test \\
\midrule
Granulometry~\cite{coenen2022learning} & Photography & 9 & 18 & 108 \\
NEU-CLS~\cite{song2013noise} & Photography & 6 & 30 & 360 \\
UHCS~\cite{decost2019high} & Microscopy & 5 & 30 & 120 \\
RIAWELC~\cite{totino2022riawelc} & X-ray & 4 & 24 & 240 \\
\bottomrule
\end{tabular}
\caption{Dataset summary. Three modalities: visible-light photography (macro and surface), optical/SEM microscopy, and X-ray radiography.}
\label{tab:datasets}
\end{table}

\textbf{Concrete Aggregate Grading (Granulometry).} 9 classes from particle size (8, 16, 32mm) $\times$ grading (coarse, medium, fine) per DIN 1045. Images are 2200$\times$3000 pixels, resized to 800px max. Ground sampling distance is computed dynamically.

\begin{figure}[H]
\centering
\includegraphics[width=0.85\textwidth]{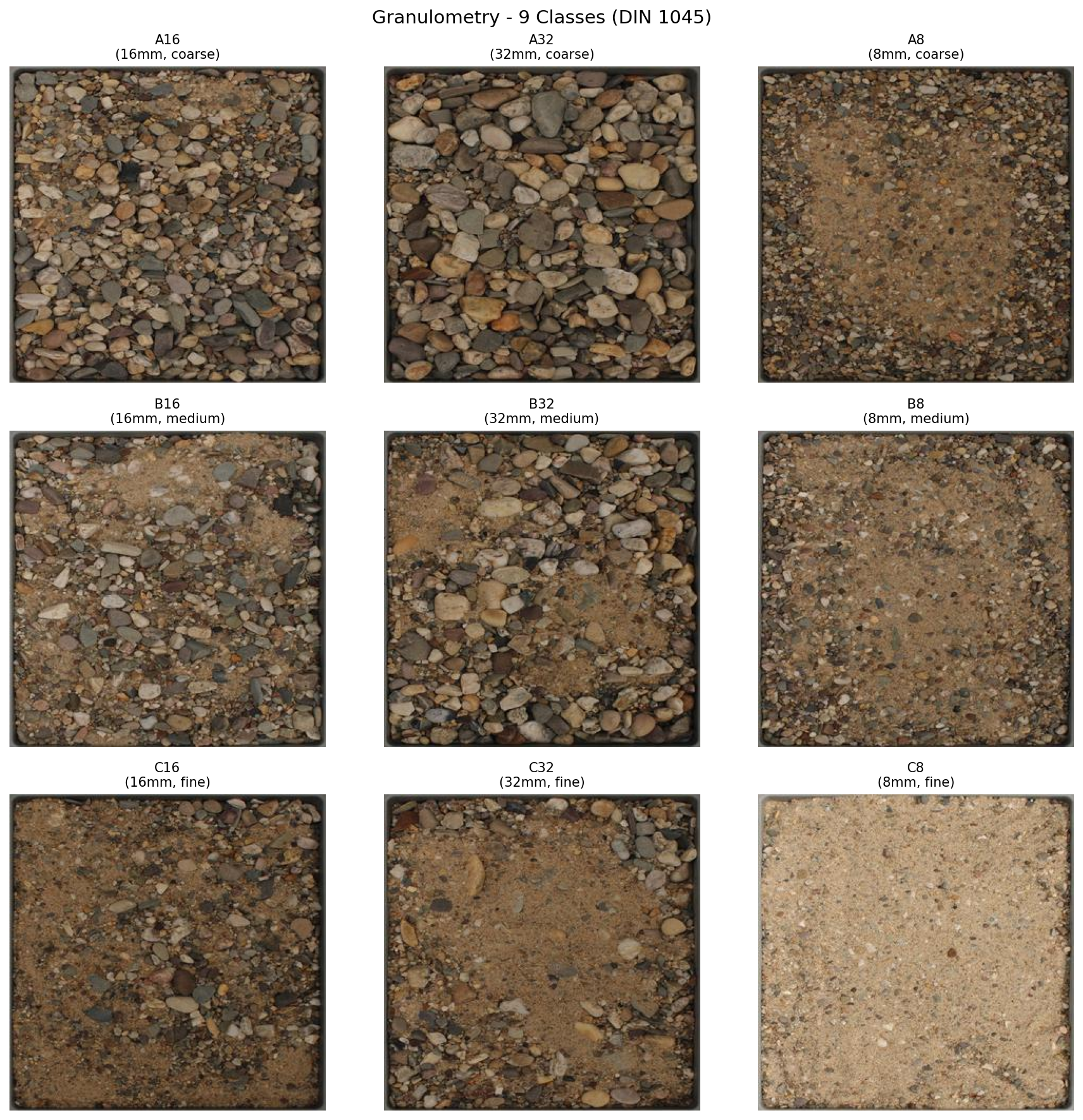}
\caption{Concrete aggregate: 9 classes (3 sizes $\times$ 3 gradings).}
\label{fig:granulometry}
\end{figure}

\textbf{Steel Surface Defects (NEU-CLS).} 6 defect classes on 200$\times$200 grayscale images of hot-rolled steel strips: crazing, inclusion, patches, pitted surface, rolled-in scale, scratches.

\begin{figure}[H]
\centering
\includegraphics[width=0.85\textwidth]{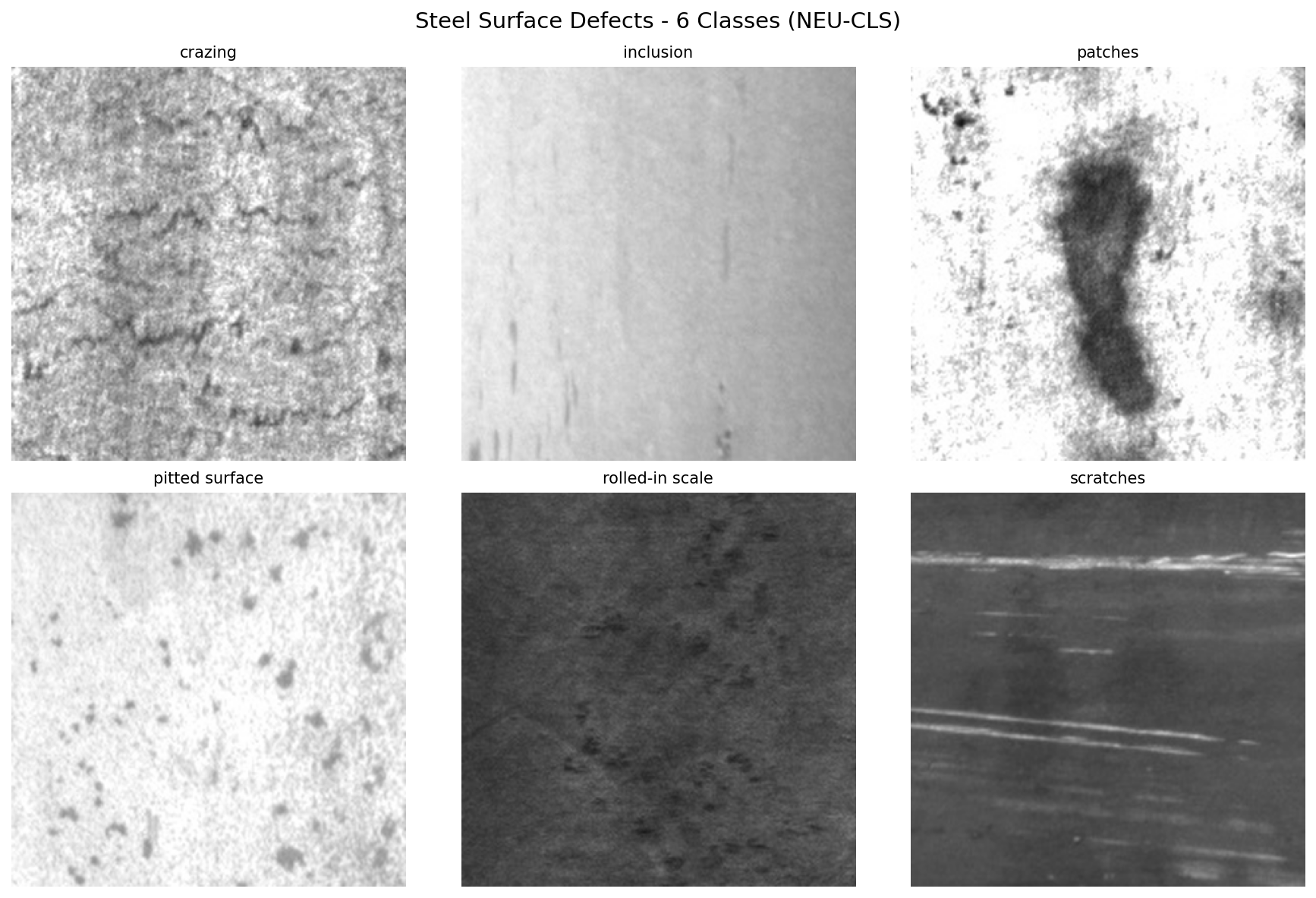}
\caption{Steel surface: 6 defect classes (NEU-CLS).}
\label{fig:steel}
\end{figure}

\textbf{Steel Microstructure (UHCS).} 5 microconstituent classes from optical/SEM micrographs: spheroidite, network, spheroidite+widmanstatten, pearlite+spheroidite, pearlite.

\begin{figure}[H]
\centering
\includegraphics[width=0.85\textwidth]{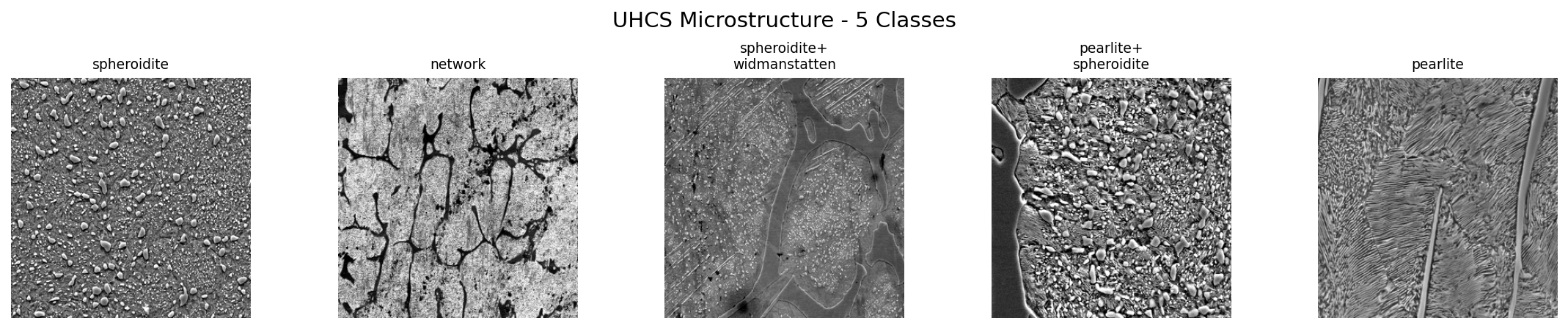}
\caption{UHCS: 5 microconstituent classes.}
\label{fig:uhcs}
\end{figure}

\textbf{Weld Defects (RIAWELC).} 4 classes on 227$\times$227 grayscale X-ray radiographs: lack of penetration, porosity, cracks, no defect.

\begin{figure}[H]
\centering
\includegraphics[width=0.85\textwidth]{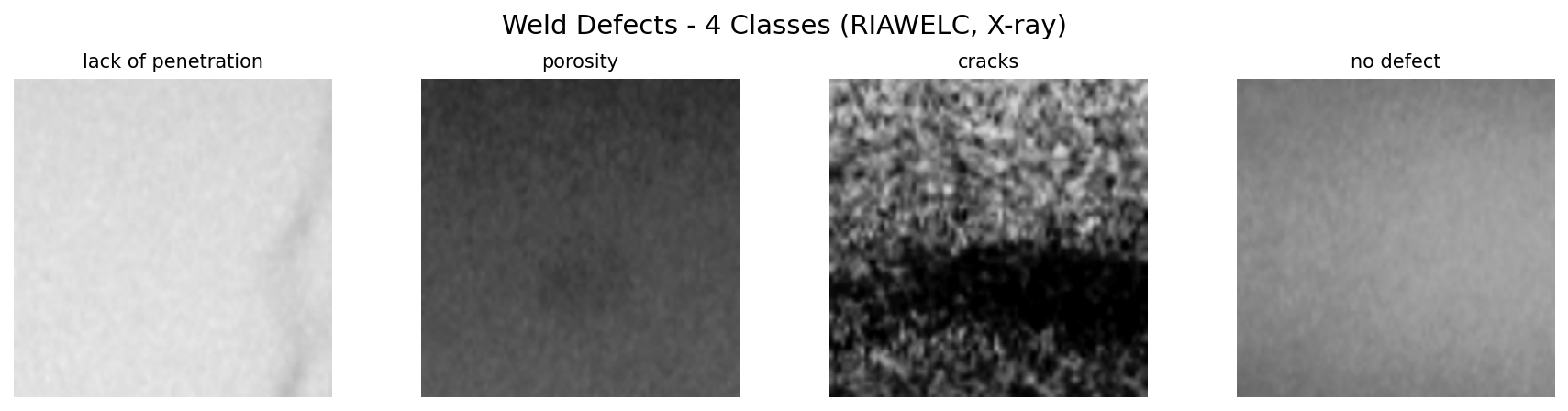}
\caption{Weld radiographs: 4 defect classes (RIAWELC).}
\label{fig:weld}
\end{figure}

\subsection{Baselines and Evaluation}

\textbf{Base model zero-shot.} Qwen2.5-VL-3B-Instruct without fine-tuning. \textbf{Frontier model few-shot.} GPT-4.1 with reference images (one per class) at inference. Note: our fine-tuned models do not receive reference images. \textbf{Direct LoRA.} Same model, same images, same hyperparameters, without reasoning augmentation.

All results are means across 4 random seeds (123, 456, 789, 1024). Each seed controls LoRA weight initialization and dropout masks. Training data, hyperparameters, and test sets are fixed. Evaluation uses temperature 0.1. For granulometry, we report ``both correct'' (size AND grading).

\section{Results}
\label{sec:results}

\subsection{Main Results}

CoT-Aug outperforms Direct LoRA on all 4 tasks across all 4 seeds (16/16 individual runs), as shown in Table~\ref{tab:main}.

\begin{table}[h]
\centering
\begin{tabular}{lcccc}
\toprule
Task & Base ZS & GPT-4.1 FS & Direct & CoT-Aug \\
\midrule
Granulometry & 12.0 & 29.6 & 75.2$\pm$1.2 & \textbf{77.8$\pm$1.5} \\
Steel Surface & 21.7 & 91.1 & 63.6$\pm$0.9 & \textbf{66.4$\pm$1.3} \\
UHCS & 60.8 & 71.7 & 64.4$\pm$0.4 & \textbf{68.8$\pm$1.1} \\
Weld Defects & 30.8 & 65.0 & 73.3$\pm$0.3 & \textbf{75.0$\pm$0.8} \\
\bottomrule
\end{tabular}
\caption{Accuracy (\%, mean$\pm$std over 4 seeds). Granulometry = both correct. Bold = best overall.}
\label{tab:main}
\end{table}

The 3B model outperforms GPT-4.1 few-shot on 3/4 tasks despite GPT-4.1 receiving reference images at inference. On weld: 75.0\% vs 65.0\% (+10.0pp). On granulometry: 77.8\% vs 29.6\% (+48.2pp). Steel is the exception where GPT-4.1's reference-image approach (91.1\%) wins. UHCS achieves $p$=0.002 (paired $t$-test); other tasks $p$=0.05--0.10 with 4 seeds.

Table~\ref{tab:seeds} shows results for each individual seed.

\begin{table}[h]
\centering\small
\begin{tabular}{l|cccc|cccc}
\toprule
& \multicolumn{4}{c|}{Direct} & \multicolumn{4}{c}{CoT-Aug} \\
Seed & Gran & Steel & UHCS & Weld & Gran & Steel & UHCS & Weld \\
\midrule
123 & 75.0 & 63.6 & 65.0 & 73.8 & 75.9 & 65.0 & 70.0 & 74.2 \\
456 & 76.9 & 63.6 & 64.2 & 73.3 & 77.8 & 66.4 & 69.2 & 74.6 \\
789 & 74.1 & 62.5 & 64.2 & 72.9 & 79.6 & 68.1 & 68.3 & 75.8 \\
1024 & 75.0 & 64.7 & 64.2 & 73.3 & 77.8 & 66.1 & 67.5 & 75.4 \\
\bottomrule
\end{tabular}
\caption{Individual seed results (\%). CoT-Aug wins all 16 comparisons.}
\label{tab:seeds}
\end{table}

\subsection{Per-Class Analysis}

Table~\ref{tab:perclass_weld} shows per-class accuracy for weld defects. Cracks is the hardest class (0\% for all baselines in zero-shot). With 6 training images, CoT-Aug reaches 50\%. Direct LoRA reaches 58\% on cracks but is worse overall because CoT-Aug gains more on no\_defect (+11pp) and lack\_of\_penetration (+7pp).

\begin{table}[h]
\centering\small
\begin{tabular}{lcc}
\toprule
Class (N=60 each) & Direct & CoT-Aug \\
\midrule
no\_defect & 87\% & \textbf{98\%} \\
porosity & 83\% & 83\% \\
lack\_of\_penetration & 65\% & \textbf{72\%} \\
cracks & \textbf{58\%} & 50\% \\
\midrule
Overall & 73.3\% & \textbf{75.8\%} \\
\bottomrule
\end{tabular}
\caption{Per-class accuracy, weld defects (seed 42).}
\label{tab:perclass_weld}
\end{table}

Table~\ref{tab:perclass_steel} shows steel surface results. The largest CoT-Aug gains are on scratches (+22pp) and patches (+23pp), where contrastive reasoning (``sharp linear grooves'' vs ``blotchy zones'') transfers well. Inclusion remains hard for both methods because dark elongated streaks are visually similar to scratches in grayscale.

\begin{table}[h]
\centering\small
\begin{tabular}{lcc}
\toprule
Class (N=60 each) & Direct & CoT-Aug \\
\midrule
pitted\_surface & 83\% & \textbf{85\%} \\
scratches & 63\% & \textbf{85\%} \\
crazing & \textbf{83\%} & 68\% \\
patches & 52\% & \textbf{75\%} \\
rolled-in\_scale & \textbf{58\%} & 55\% \\
inclusion & \textbf{38\%} & 32\% \\
\midrule
Overall & 63.1\% & \textbf{66.7\%} \\
\bottomrule
\end{tabular}
\caption{Per-class accuracy, steel surface (seed 42).}
\label{tab:perclass_steel}
\end{table}

For granulometry (9 classes), size classification is easier than grading: 91.7\% vs 86.1\% for CoT-Aug. The hardest class is C32 (32mm, fine grading) at 42\% both-correct. This is an unusual combination (large stones with fine distribution) that may be underrepresented in 2 training images.

\subsection{Improvement Is From Reasoning Quality, Not Data Volume}

CoT-Aug uses $4N$ examples while Direct uses $N$. To isolate reasoning quality from training budget, we trained Direct-4$\times$: each image duplicated 4 times, matching CoT-Aug's steps exactly (Table~\ref{tab:budget}).

\begin{table}[h]
\centering
\begin{tabular}{lccc}
\toprule
Task & Direct & Direct-4$\times$ & CoT-Aug \\
\midrule
Granulometry & 75.2 & 74.1 & \textbf{77.8} \\
Steel Surface & 63.6 & 63.1 & \textbf{66.4} \\
UHCS & 64.4 & \textbf{70.8} & 68.8 \\
Weld Defects & 73.3 & 72.9 & \textbf{75.0} \\
\bottomrule
\end{tabular}
\caption{Equal-budget control (\%). Direct-4$\times$ matches CoT-Aug's optimizer steps.}
\label{tab:budget}
\end{table}

On 3/4 tasks, Direct-4$\times$ performs at or below Direct despite 4$\times$ more steps. The training loss tells the story: Direct-4$\times$ reaches near-zero loss by epoch 30 (complete memorization of repeated JSON targets), while CoT-Aug converges to 0.02--0.05 (still learning from diverse descriptions at epoch 40). Naive repetition causes overfitting; diverse reasoning provides genuine signal.

UHCS is the exception: compound class names (``spheroidite+widmanstatten'') are 3--4 tokens long. Repeated exposure helps the model memorize these unusual label sequences.

\subsection{Answer-Conditioning Is Essential}

We tested what happens without answer-conditioning. Gemini 2.5 Pro classified each training image freely (no correct answer provided). We trained on its predictions even when wrong. Table~\ref{tab:uncond} shows results alongside Gemini's measured accuracy.

\begin{table}[h]
\centering
\begin{tabular}{lccccc}
\toprule
Task & Gemini ZS & Gemini FS & Uncond. & Direct & Cond. \\
\midrule
Granulometry & 24.1 & 27.8 & 57.4 & 75.2 & \textbf{77.8} \\
Weld & 63.7 & 62.1 & 73.3 & 73.3 & \textbf{75.0} \\
\bottomrule
\end{tabular}
\caption{Unconditioned CoT (\%). Gemini 2.5 Pro as teacher without knowing the correct answer.}
\label{tab:uncond}
\end{table}

The pattern is clear:
\begin{itemize}
\item \textbf{Teacher accuracy 24\% (granulometry):} Unconditioned CoT drops 17.8pp below Direct. Wrong reasoning poisons the model.
\item \textbf{Teacher accuracy 63.7\% (weld):} Unconditioned CoT matches Direct exactly. Correct and incorrect reasoning cancel out.
\item \textbf{Teacher accuracy 100\% (conditioned):} CoT improves over Direct by +1.7 to +4.4pp.
\end{itemize}

We used Gemini 2.5 Pro (a model stronger than GPT-4.1 on most vision benchmarks) to demonstrate this principle is model-independent. If even a state-of-the-art frontier model produces harmful training data without answer-conditioning, the design choice is essential, not optional.

\subsection{Training Convergence}

Direct models start at low loss (0.17--0.19) since targets are short JSON strings (10--30 tokens). They converge to near-zero by epoch 25 (complete memorization). CoT-augmented models start at higher loss (1.35--1.55) because targets are 50--150 tokens. They converge smoothly to 0.02--0.05 by epoch 40.

\begin{table}[h]
\centering\small
\begin{tabular}{l|ccc|ccc}
\toprule
& \multicolumn{3}{c|}{Direct Loss} & \multicolumn{3}{c}{CoT-Aug Loss} \\
Task & Ep.1 & Ep.20 & Ep.40 & Ep.1 & Ep.20 & Ep.40 \\
\midrule
Granulometry & 0.184 & 0.084 & 0.023 & 1.396 & 0.131 & 0.046 \\
Weld & 0.174 & 0.006 & 0.000 & 1.351 & 0.113 & 0.022 \\
\bottomrule
\end{tabular}
\caption{Training loss at epochs 1, 20, 40. Direct reaches near-zero (memorization); CoT-Aug retains learning signal throughout.}
\label{tab:convergence}
\end{table}

The weld Direct model reaching near-zero loss by epoch 25 indicates complete memorization of 24 JSON outputs. This motivates the augmented approach: by providing diverse descriptions, the model has more to learn from and is less prone to overfitting.

\section{Discussion}
\label{sec:discussion}

\textbf{When does CoT distillation help most?} The gain correlates with task complexity. Granulometry (9 classes, relational reasoning about gap patterns) and UHCS (compound classes requiring co-feature detection) show the largest gains. Weld (4 well-separated classes) shows smaller but consistent gain. The contrastive descriptions are most valuable where classes are visually confusable.

\textbf{Characteristic confusion patterns.} Each task has specific confusion pairs that CoT addresses:
\begin{itemize}
\item Granulometry: coarse vs medium at small sizes (subtle gap differences)
\item Steel: inclusion $\leftrightarrow$ scratches (both dark elongated features)
\item UHCS: spheroidite+widmanstatten $\rightarrow$ spheroidite (requires detecting co-existing features)
\item Weld: cracks $\leftrightarrow$ lack\_of\_penetration (both dark lines in radiographs)
\end{itemize}
CoT descriptions explicitly teach these distinctions (``circular spots, NOT jagged lines'') which is information that JSON-only training cannot convey.

\textbf{The role of answer-conditioning.} Our unconditioned experiment reveals a clear relationship: performance tracks teacher accuracy. At 24\% teacher accuracy, unconditioned CoT is catastrophic. At 63.7\%, neutral. At 100\% (conditioned), positive. This makes answer-conditioning essential for any domain where frontier models lack expertise, which is precisely the setting where the method is most needed.

\textbf{Prompt design matters more than expected.} The 18.1pp drop from using weaker class definitions (same images, same hyperparameters) suggests that prompt engineering is not auxiliary to the method. It is the method. The class definitions are what the frontier model uses to generate structured reasoning. Weak definitions produce weak reasoning.

\textbf{Practical deployment.} The complete pipeline (CoT generation via API + LoRA training) completes in under 2 hours per task on a single NVIDIA L4 24GB GPU (\$1/hr). Training times range from 19 minutes (granulometry Direct) to 111 minutes (UHCS CoT-Aug). A factory engineer with 18--30 labeled images can deploy a domain-adapted VLM the same day.

\textbf{Limitations.} All experiments use Qwen2.5-VL-3B. We do not test other architectures or model sizes. We use 3 CoT descriptions per image; the optimal number may vary. Steel surface remains a case where reference-image approaches outperform. The GPT-4.1 few-shot comparison is not apples-to-apples (it receives reference images at inference; our model does not).

\section{Conclusion}
\label{sec:conclusion}

We presented answer-conditioned CoT distillation for adapting small VLMs to industrial inspection with minimal labeled data. Across 4 tasks, 3 modalities, and 32 training runs, the method consistently outperforms direct fine-tuning (16/16 wins). The gain comes from reasoning quality (not training budget), and answer-conditioning is essential (without it, performance drops 17.8pp). A 3B model fine-tuned on 24 images outperforms both GPT-4.1 and Gemini 2.5 Pro on specialized tasks. The full pipeline runs in under 2 hours on commodity hardware.

\bibliographystyle{plainnat}
\bibliography{references}

\begin{thebibliography}{19}
\providecommand{\natexlab}[1]{#1}
\providecommand{\url}[1]{\texttt{#1}}
\expandafter\ifx\csname urlstyle\endcsname\relax
  \providecommand{\doi}[1]{doi: #1}\else
  \providecommand{\doi}{doi: \begingroup \urlstyle{rm}\Url}\fi

\bibitem[Coenen et~al.(2022)Coenen, Beyer, Heipke, and
  Haist]{coenen2022learning}
M.~Coenen, D.~Beyer, C.~Heipke, and M.~Haist.
\newblock Learning to sieve: Prediction of grading curves from images of
  concrete aggregate.
\newblock \emph{ISPRS Annals of Photogrammetry, Remote Sensing and Spatial
  Information Sciences}, V-2-2022:\penalty0 227--235, 2022.

\bibitem[Czimmermann et~al.(2020)Czimmermann, Ciuti, Milazzo, Chiurazzi,
  Roccella, Oddo, and Dario]{czimmermann2020visual}
T.~Czimmermann, G.~Ciuti, M.~Milazzo, M.~Chiurazzi, S.~Roccella, C.M. Oddo, and
  P.~Dario.
\newblock Visual-based defect detection and classification approaches for
  industrial applications---a survey.
\newblock \emph{Sensors}, 20\penalty0 (5):\penalty0 1459, 2020.

\bibitem[DeCost et~al.(2019)DeCost, Lei, Francis, and Holm]{decost2019high}
B.L. DeCost, B.~Lei, T.~Francis, and E.A. Holm.
\newblock High throughput quantitative metallography for complex
  microstructures using deep learning: A case study in ultrahigh carbon steel.
\newblock \emph{Microscopy and Microanalysis}, 25\penalty0 (1):\penalty0
  21--29, 2019.

\bibitem[{German Institute for Standardization (Deutsches Institut f{\"u}r
  Normung)}(2008)]{din1045}
{German Institute for Standardization (Deutsches Institut f{\"u}r Normung)}.
\newblock {DIN} 1045-2:2008. {C}oncrete, reinforced and prestressed concrete
  structures, 2008.

\bibitem[Hsieh et~al.(2023)Hsieh, Li, Yeh, Nakhost, Fujii, Ratner, Krishna,
  Lee, and Pfister]{hsieh2023distilling}
C.Y. Hsieh, C.L. Li, C.K. Yeh, H.~Nakhost, Y.~Fujii, A.~Ratner, R.~Krishna,
  C.Y. Lee, and T.~Pfister.
\newblock Distilling step-by-step! outperforming larger language models with
  less training data and smaller model sizes.
\newblock In \emph{Findings of ACL}, 2023.

\bibitem[Hu et~al.(2021)Hu, Shen, Wallis, Allen-Zhu, Li, Wang, Wang, and
  Chen]{hu2021lora}
E.J. Hu, Y.~Shen, P.~Wallis, Z.~Allen-Zhu, Y.~Li, S.~Wang, L.~Wang, and
  W.~Chen.
\newblock {LoRA}: Low-rank adaptation of large language models.
\newblock \emph{arXiv preprint arXiv:2106.09685}, 2021.

\bibitem[Kang et~al.(2025)Kang, Aljundi, Dorovatas, and
  Alahari]{online2025incontext}
Z.~Kang, R.~Aljundi, V.~Dorovatas, and K.~Alahari.
\newblock Online in-context distillation for low-resource vision language
  models.
\newblock \emph{arXiv preprint arXiv:2510.18117}, 2025.

\bibitem[Lee et~al.(2024)Lee, Hachiuma, Wang, Ro, and Wu]{vlsi2024verbalized}
B.K. Lee, R.~Hachiuma, Y.C.F. Wang, Y.M. Ro, and Y.H. Wu.
\newblock {VLsI}: Verbalized layers-to-interactions from large to small vision
  language models.
\newblock \emph{arXiv preprint arXiv:2412.01822}, 2024.

\bibitem[Megahed et~al.(2025)Megahed, Chen, Colosimo, Grasso, Jones-Farmer,
  Knoth, Sun, and Zwetsloot]{clip2025fewshot}
F.M. Megahed, Y.J. Chen, B.M. Colosimo, M.L.G. Grasso, L.A. Jones-Farmer,
  S.~Knoth, H.~Sun, and I.~Zwetsloot.
\newblock Adapting {OpenAI}'s {CLIP} model for few-shot image inspection in
  manufacturing quality control.
\newblock \emph{arXiv preprint arXiv:2501.12596}, 2025.

\bibitem[Moenck et~al.(2024)Moenck, Thieu, Koch, and
  Sch{\"u}ppstuhl]{adapting2024industry}
K.~Moenck, D.T. Thieu, J.~Koch, and T.~Sch{\"u}ppstuhl.
\newblock Industrial language-image dataset ({ILID}): Adapting vision
  foundation models for industrial settings.
\newblock \emph{arXiv preprint arXiv:2406.09637}, 2024.

\bibitem[Rajani et~al.(2019)Rajani, McCann, Xiong, and
  Socher]{rajani2019explain}
Nazneen~Fatema Rajani, Bryan McCann, Caiming Xiong, and Richard Socher.
\newblock Explain yourself! leveraging language models for commonsense
  reasoning.
\newblock In \emph{ACL}, 2019.

\bibitem[Song and Yan(2013)]{song2013noise}
K.~Song and Y.~Yan.
\newblock A noise robust method based on completed local binary patterns for
  hot-rolled steel strip surface defects.
\newblock \emph{Applied Surface Science}, 285:\penalty0 858--864, 2013.

\bibitem[Totino et~al.(2022)Totino, Spagnolo, and Perri]{totino2022riawelc}
B.~Totino, F.~Spagnolo, and S.~Perri.
\newblock {RIAWELC}: A novel dataset of radiographic images for automatic weld
  defects classification.
\newblock \emph{Research Square (Preprint)}, 2022.

\bibitem[Wang et~al.(2024)Wang, Bai, Tan, Wang, Fan, Bai, Chen, Liu, Wang, Ge,
  Fan, Dang, Du, Ren, Men, Liu, Zhou, Zhou, and Lin]{wang2024qwen2vl}
P.~Wang, S.~Bai, S.~Tan, S.~Wang, Z.~Fan, J.~Bai, K.~Chen, X.~Liu, J.~Wang,
  W.~Ge, Y.~Fan, K.~Dang, M.~Du, X.~Ren, R.~Men, D.~Liu, C.~Zhou, J.~Zhou, and
  J.~Lin.
\newblock Qwen2-{VL}: Enhancing vision-language model's perception of the world
  at any resolution.
\newblock \emph{arXiv preprint arXiv:2409.12191}, 2024.

\bibitem[Wei et~al.(2022)Wei, Wang, Schuurmans, Bosma, Ichter, Xia, Chi, Le,
  and Zhou]{wei2022chain}
J.~Wei, X.~Wang, D.~Schuurmans, M.~Bosma, B.~Ichter, F.~Xia, E.~Chi, Q.V. Le,
  and D.~Zhou.
\newblock Chain-of-thought prompting elicits reasoning in large language
  models.
\newblock In \emph{Advances in Neural Information Processing Systems},
  volume~35, pages 24824--24837, 2022.

\bibitem[Xia et~al.(2025)Xia, Tong, Zang, Shao, and Zhou]{xia2025bootstrapping}
Jiaer Xia, Bingkui Tong, Yuhang Zang, Rui Shao, and Kaiyang Zhou.
\newblock Bootstrapping grounded chain-of-thought in multimodal {LLMs} for
  data-efficient model adaptation.
\newblock In \emph{ICCV}, 2025.
\newblock arXiv:2507.02859.

\bibitem[Zelikman et~al.(2022)Zelikman, Wu, Mu, and Goodman]{zelikman2022star}
Eric Zelikman, Yuhuai Wu, Jesse Mu, and Noah Goodman.
\newblock {STaR}: Bootstrapping reasoning with reasoning.
\newblock In \emph{NeurIPS}, 2022.

\bibitem[Zhang et~al.(2024)Zhang, Meyer, Lu, Shrivastava, and
  Ravichandran]{kd2024longtail}
Z.~Zhang, G.P. Meyer, Z.~Lu, A.~Shrivastava, and A.~Ravichandran.
\newblock Knowledge distillation from {VLM} for long-tail visual recognition.
\newblock \emph{arXiv preprint arXiv:2408.16930}, 2024.

\bibitem[Zohar et~al.(2025)Zohar, Wang, Bitton, Szpektor, and
  Yeung-Levy]{zohar2025videostar}
Orr Zohar, Xiaohan Wang, Yonatan Bitton, Idan Szpektor, and Serena Yeung-Levy.
\newblock {Video-STaR}: Self-training enables video instruction tuning with any
  supervision.
\newblock In \emph{ICLR}, 2025.

\end{thebibliography}

\end{document}